**Leveraging the Capabilities of Connected and Autonomous Vehicles and Multi-Agent Reinforcement Learning to Mitigate Highway Bottleneck Congestion**

**Paul (Young Joun) Ha**
Graduate Research Assistant, Center for Connected and Automated Transportation (CCAT), and Lyles School of Civil Engineering, Purdue University, West Lafayette, IN, 47907.
Email: ha55@purdue.edu
ORCID #: 0000-0002-8511-8010

**Sikai Chen\***
Postdoctoral Research Fellow, Center for Connected and Automated Transportation (CCAT), and Lyles School of Civil Engineering, Purdue University, West Lafayette, IN, 47907.
Email: chen1670@purdue.edu; and
Visiting Research Fellow, Robotics Institute, School of Computer Science, Carnegie Mellon University, Pittsburgh, PA, 15213.
Email: sikaichen@cmu.edu
ORCID #: 0000-0002-5931-5619
(Corresponding author)

**Jiqian Dong**
Graduate Research Assistant, Center for Connected and Automated Transportation (CCAT), and Lyles School of Civil Engineering, Purdue University, West Lafayette, IN, 47907.
Email: dong282@purdue.edu
ORCID #: 0000-0002-2924-5728

**Runjia Du**
Graduate Research Assistant, Center for Connected and Automated Transportation (CCAT), and Lyles School of Civil Engineering, Purdue University, West Lafayette, IN, 47907.
Email: du187@purdue.edu
ORCID #: 0000-0001-8403-4715

**Yujie Li**
Graduate Research Assistant, Center for Connected and Automated Transportation (CCAT), and Lyles School of Civil Engineering, Purdue University, West Lafayette, IN, 47907.
Email: li2804@purdue.edu
ORCID #: 0000-0002-0656-4603

**Samuel Labi**
Professor, Center for Connected and Automated Transportation (CCAT), and Lyles School of Civil Engineering, Purdue University, West Lafayette, IN, 47907.
Email: labi@purdue.edu
ORCID #: 0000-0001-9830-2071

*Ha, Chen, Dong, Du, Li, and Labi*

**ABSTRACT**

Highway bottleneck refers to the localized constriction of traffic flow in a given corridor when the traffic demand exceeds its available capacity. Active Traffic Management strategies are often adopted in real-time to address such sudden flow breakdowns. When queuing is imminent, Speed Harmonization (SH), which adjusts speeds in upstream traffic to mitigate traffic showckwaves downstream, can be applied. However, because SH depends on driver awareness and compliance, it may not always be effective in mitigating congestion. The use of multiagent reinforcement learning for collaborative learning, is a promising solution to this challenge. By incorporating this technique in the control algorithms of connected and autonomous vehicle (CAV), it may be possible to train the CAVs to make joint decisions that can mitigate highway bottleneck congestion without human driver compliance to altered speed limits. In this regard, we present an RL-based multi-agent CAV control model to operate in mixed traffic (both CAVs and human-driven vehicles (HDVs)). The results suggest that even at CAV percent share of corridor traffic as low as 10%, CAVs can significantly mitigate bottlenecks in highway traffic. Another objective was to assess the efficacy of the RL-based controller vis-à-vis that of the rule-based controller. In addressing this objective, we duly recognize that one of the main challenges of RL-based CAV controllers is the variety and complexity of inputs that exist in the real world, such as the information provided to the CAV by other connected entities and sensed information. These translate as dynamic length inputs which are difficult to process and learn from. For this reason, we propose the use of Graphical Convolution Networks (GCN), a specific RL technique, to preserve information network topology and corresponding dynamic length inputs. We then use this, combined with Deep Deterministic Policy Gradient (DDPG), to carry out multi-agent training for congestion mitigation using the CAV controllers. From the results of the simulation experiments, we find that for purposes of bottleneck congestion mitigation, the RL-based controller has superior performance compared with the rule-based controller.

**INTRODUCTION AND BACKGROUND**

Highway infrastructure features such as ramps, merges, lane drops (sudden reduction in the number of lanes along a corridor), and tollbooths are designed to operate efficiently without much delay for a specific traffic demand. However, inflows that exceed the design capacity can create operationally influenced deficiencies, resulting in anomalous traffic flow conditions such as localized constriction of the traffic flow, also referred to as bottlenecks (FHWA, 2011). Bottlenecks also arise from unforeseen events such as stalled vehicles on the roadway due to work zones, crashes, lack of fuel, or roadway debris downstream that disables at least one lane, leaving a reduced number of lanes for the traffic. This situation leads to abrupt changes from free-flowing traffic conditions to congested traffic conditions (Persaud et al., 1998), which impairs travel efficiency, increases the changes of secondary incidents, increases fuel consumption and emissions, and exacerbates driver frustration and discomfort (Abdel-Aty et al., 2006; Barth and Boriboonsomsin, 2009, 2008; EPA, 2015; Li et al., 2020a; Figliozzi, 2010; Lee et al., 2006). The Environmental Protection Agency reports that the transportation sector is responsible for almost 30% of total US greenhouse gas emissions (EPA, 2019). Similarly, heavy congestion is a strong contributing factor to traffic accidents (Retallack and Ostendorf, 2019). The National Highway Traffic Safety Administration (NHTSA) reports that over 35,000 traffic crash fatalities occurred in 2018 (NHTSA, 2018) and a significant fraction of these crashes can be attributed directly or indirectly to highway congestion. Clearly, the socioeconomic repercussions of traffic congestion are significant and highlight the need to mitigate congestion wherever possible.

Further, highway bottlenecks can cause rapid oscillations in vehicle speeds and densities, which can rapidly propagate upstream (Sugiyamal et al., 2008). In other words, the extended consequences of highway bottlenecks (environmental, economic, and safety) are not contained only in localized areas where the bottleneck occurs but over a far wider stretch, rather, exposing more travelers to the risks associated with congestion. Thus, numerous researchers have studied both the nature and prediction of flow breakdown as well as mitigation strategies to alleviate or prevent congestion. The various approaches that have been considered for traffic flow breakdown mitigation include preemptive lane change maneuvers (Nagalur Subraveti et al., 2019) as well as speed harmonization (SH), which reduces the speed variance of the vehicles in an area of interest (Abdel-Aty et al., 2006; Dowling et al., 2016; Lee et al., 2006; Mahmassani, 2016; Malikopoulos et al., 2019; Strömgren and Lind, 2016). However, such flow breakdown mitigation strategies often rely on high levels of driver compliance, rendering flow breakdown mitigation to be dependent on the awareness and willingness of human drivers. Fortunately, recent advancements in vehicle automation and connectivity technology has shown that it has significant potential to enhance the mobility, safety, and emissions of the traditional transportation network (Auld et al., 2017; Gao et al., 2016; Ha, 2019; Chen, 2019; Ilgin Guler et al., 2014; Mahmassani, 2014; Ha et al., 2020; Chen et al., 2020; Dong et al., 2020; Du et al., 2020). Additionally, connected and autonomous vehicles (CAVs) can be utilized for SH in various congested situations (Ma et al., 2016; Tajalli and Hajbabaie, 2018; Wang et al., 2016; Li et al., 2020b). Therefore, this study compares the use of a deep reinforcement-learning (DRL) based CAV controller against a rule-based vehicle controller which performs predefined maneuvers in traffic to determine the potential for mitigating highway bottleneck congestion using CAVs.

**Active Traffic Management**

Active Traffic Management (ATM) is a broad term that refers to various practical strategies to mitigate congestion and assuage flow conditions (Mohammad Mirshahi et al., 2007). Speed Harmonization (SH), a specific type of ATM, refers to the control of the spatial-temporal oscillations in vehicle speeds. The benefits of SH is well-studied: it improves mobility and safety, and reduces emissions (Ma et al., 2016; Strömgren and Lind, 2016; Talebpour et al., 2013a).

To achieve speed harmonization, many researchers suggest variable speed limits (VSL), which alters the upstream speed limits based on the predicted or prevailing flow breakdown in downstream traffic. The fundamental idea behind this strategy is to reduce incident-related risks and to keep operational performance above a minimum desirable level to promote traffic stability (Filipovska et al., 2019; Lu and Shladover, 2014; Ma et al., 2016). A considerable number of studies have evaluated the efficacy of VSL schemes, in the context of traditional traffic streams as well as with intelligent systems (Abdel-Aty et al., 2006; Dowling et al., 2016; Ghiasi et al., 2017; Lee et al., 2006; Ma et al., 2016; Mahmassani, 2016; Malikopoulos et al., 2019; Strömgren and Lind, 2016; Talebpour et al., 2013b). Lu et al. reviewed various VSL studies and advisories, and showed that it can improve traffic safety and flow conditions (Lu and Shladover, 2014). In a study that focused on safety impacts of VSL, Lee et al. showed that VSL may reduce crash potential by 5–17% (Lee et al., 2006). Filipovska et al. presented approaches to not only predict but also to mitigate flow breakdown using VSL for traffic network prone to inclement weather, and found that VSL designed for such conditions can mitigate flow breakdown (Filipovska et al., 2019). Clearly, VSL can be used to achieve speed harmonization. However, their successful deployment depends on the compliance of human drivers in the traffic stream. If drivers do not comply with VSL, implementation becomes unsuccessful, and flow breakdown may not be mitigated. Due to the enforcement challenges, SH strategies that require human driver compliance, such as VSL, has often been criticized as an unrealistic flow breakdown mitigation strategy.

Fortunately, the emergence of connected and autonomous vehicles (CAVs) has offered, as an alternative to VSL and other traditional SH technologies, an unprecedented and potentially efficacious tool to achieve traffic speed harmonization. CAVs, in their role as a probe for real-time, disaggregate data collection or as a controller for the implementation and encouragement of flow breakdown mitigation strategies, have been investigated in past research. In one such study, researchers combined intelligent transportation systems with SH, and developed an algorithm to utilize vehicle-to-infrastructure communication to determine the set speed for an adaptive cruise control system or variable message signs (Lu et al., 2015). Another study by Talebpour et al., observed that speed harmonization strategies based on connected vehicles can mitigate traffic flow breakdown and improve safety at low market penetrations (Talebpour et al., 2013a). Similarly, Ghiasi et al. presented an algorithm for trajectory-smoothing using a single connected vehicle to mitigate flow breakdown in single-lane environments (Ghiasi et al., 2017). These studies provide evidence of past researchers regarding the leveraging potential of vehicular connectivity and automation for traffic speed harmonization. In a bid to build further on this momentum, this study seeks to use a centralized DRL-based CAV controller that can effectively adjust the velocities of relevant agents prior to encountering the bottleneck-induced capacity reduction, for a traffic stream comprised of CAVs and HDVs. The goal is to mitigate the downstream flow breakdown and avoiding or reducing the accompanying congestion.

The remainder of this paper is organized as follows: first, we discuss the contributions of the paper, then we present the study methodology which includes the model architecture and the reinforcement learning algorithm. We then discuss the experiment settings for training and simulation not only of the proposed controller but also of a standard controller based on the traditional car-following model. Finally, we discuss the results and the implications on the practice.

**Contributions of the Paper**

From the foregoing discussion, the intended contributions of this paper are: proposition of a novel information fusion method (GCN) to aggregate proximal and distant information from onboard sensors and vehicular connectivity, and the use of several CAVs to mitigate highway bottleneck congestion in free-flowing (i.e., no constraints in vehicle lateral and longitudinal control, in other words, the vehicle is allowed to accelerate, decelerate or make lane changes) in a mixed traffic stream comprised of both CAVs and HDVs.

**METHODOLOGY**

**Reinforcement Learning and Graphical Neural Networks**

This study involves the use of reinforcement learning (RL) to train CAV agents. In RL settings, agents explore the environment and are rewarded for desirable actions and penalized (or simply not rewarded) for undesirable actions (Lillicrap et al., 2016; Dong et al., 2020). This feedback system of actions, states, and rewards ultimately allows for the agent to tune their parameters to perform actions that maximize their final reward. For this study, RL is combined with Graphical Convolution to reward congestion mitigating actions while penalizing actions that may perpetuate congestion in the highway bottleneck environment.

One way to model the information flow path and decision dependency is the use of a graphical approach. By presenting objects of interest as nodes (in this case, CAVs) and the information connection as links, the system of interconnected vehicles in the training environment can be represented as a graph. The CAVs trained for this study are connected with each other and transmit locally sensed HDV information to other CAVs, thereby creating a graphical reconstruction of the traffic stream. Figure 1 below shows a simple graphical illustration of the vehicular network comprised of 3 CAVs and their information flow. Each CAV possesses a defined sensing range, and information of HDVs within the range are collected and shared, ultimately leading to joint decisions.

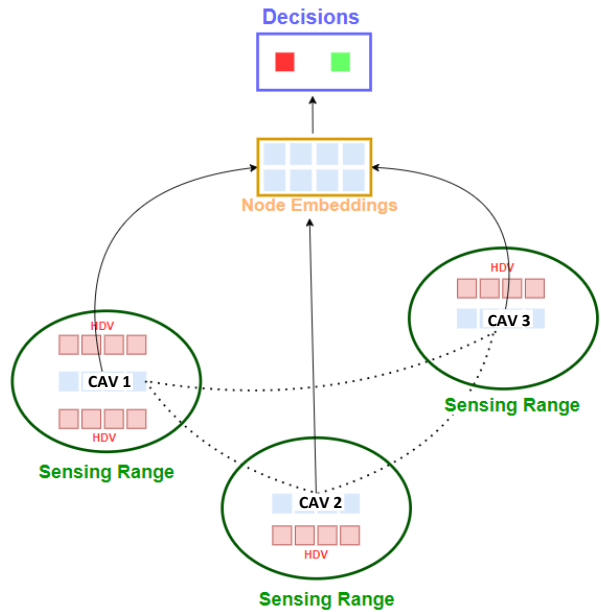

**Figure 1: Vehicular Network Information Topology**

Graph Neural Networks (GNN), which are deep learning models that preserve the dependence of graphs using nodes and edges (Zhou et al., 2018), are an increasingly popular approach used to construct neural networks with a wide range of applications (Derrible and Kennedy, 2011; Kipf and Welling, 2019; Scarselli et al., 2009; Dong et al., 2020b) due to their high interpretability and performance. One of the main advantages of GNN is the ability to construct node embeddings from relational data representations on both node features as well as neighboring node features. Therefore, it is highly useful in transportation applications where trained vehicles are surrounded by a dynamic number of observable entities.

To make maneuvering decisions, CAVs use both sensed and shared information about the traffic environment. As such, data from both sensed and shared sources must be fused in order to properly and holistically characterize the prevailing traffic conditions. Adopting the Graphical Convolutional Network (GCN) enables information from small clusters of nodes to become aggregated to generate node embeddings (Kipf and Welling, 2019). Therefore, the information fusion required in this study can be adequately performed using GCN.

Figure 2 presents the model architecture used in this study. It consists of a CAV network, which was shown in detail in Figure 1, which collects and shares operational information (such as positions and velocities) from the vehicular network information as node features into an FCN encoder, and the consequent node embeddings are the feature matrix inputted into the GCN block. Further, the adjacency matrix presents the network topology for data flow and decision dependency, which also serves as an input to the GCN block. The weights of the GCN layers can assist CAVs to focus on relevant information depending on short- or long-term goals. As this study considers both immediate lane change decisions as well as avoidance of downstream congestion, the ability to focus on relevant information has great merit. Then, the GCN block outputs the feature embedding map which is the fusion of the dynamic length inputs. These node embeddings are then fed into the actor-critic model, shown in Figure 3, to perform collaborative driving maneuvers between all agents.

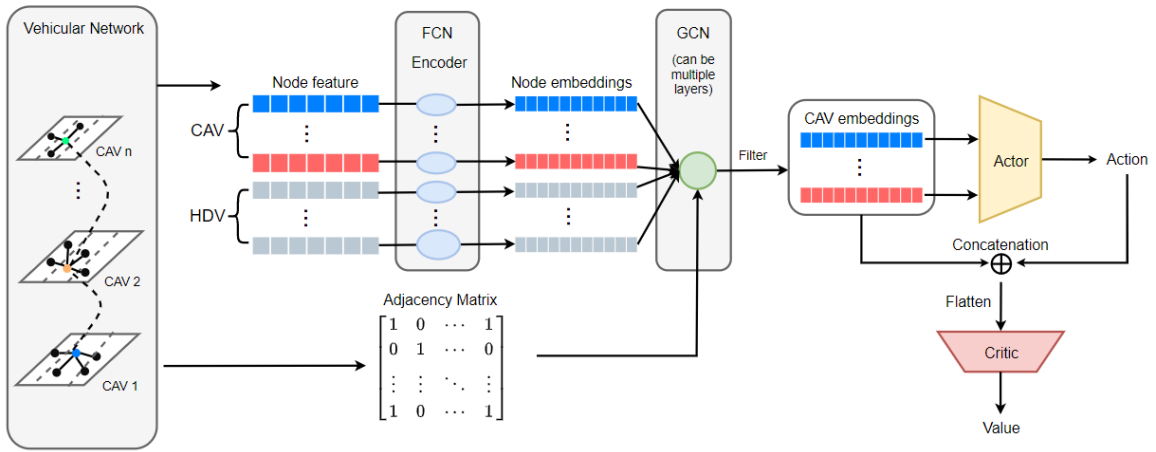

**Figure 2: Model Architecture**

The goal of the Graphical Convolution Network is to learn a function of features on a graph $\mathcal{G} = (\mathcal{V}, \mathcal{E})$ which takes the following inputs: a feature description $x_i$ for each node $i$ summarized in an $N \times D$ feature matrix $X$, where $N$ is the number of nodes and $D$ is the dimension of input features; and a representative description of the graph structure in matrix form, typically in the form of an adjacency matrix $A$ (Kipf and Welling, 2019; Tungnd, n.d.). The output is an $N \times F$ feature matrix, denoted $Z$. The graphic convolution can be applied as follows:

$$f(H^{(l)}, A) = \sigma(AH^{(l)}W^{(l)})$$

where $W^{(l)}$ is a weight matrix for the $l^{th}$ neural network layer and $\sigma(\cdot)$ is a non-linear activation function such as ReLU. However, in this formulation, the output feature map only consists of the neighboring node information. As such, the adjacency matrix requires a self-loop prior to convolution. Additionally, matrix $A$ is normalized using $D^{-1/2}AD^{-1/2}$, where $D$ is the diagonal node degree matrix. Therefore, the computation rule for graph convolution is formulated as:

$$f(H^{(l)}, A) = \sigma(\widehat{D}^{-1/2}\hat{A}\widehat{D}^{-1/2}H^{(l)}W^{(l)})$$

where $\hat{A} = A + I$ and $\widehat{D}$ is the diagonal node degree matrix of $\hat{A}$.

To achieve speed harmonization using CAVs, the decision processor on a given CAV requires both proximal information of vehicles in its surroundings (local information) as well as distant information of vehicles further downstream (global information). The local information allows a given CAV to constrain its short-term decisions to be feasible (i.e., immediate lane-change maneuvers). On the other hand, global information is useful for a given CAV to make long-term decisions for planning (i.e., routing, lane occupancy, exit intentions, etc.). Therefore, it follows that a safe and efficient CAV, in attempts to mitigate highway bottleneck congestion, must learn to utilize both local and global information.

## DDPG Agent

This study utilizes the Deep Deterministic Policy Gradient (DDPG), presented by Lillicrap et al. (Lillicrap et al., 2016), as the reinforcement learning agent. DDPG is closely related to Deep Q-Learning (DQL), but it is applied to continuous actions. As such, DDPG is an off-policy algorithm with an actor-critic network structure that endeavors to converge as:

$$Q \cong r + d \cdot Q_{next}$$

where $r$ is the cumulative reward and $Q_{next}$ is the Q-value for the next time step.

Figure 3 shows an illustration of the training process for the $Q$ and $Q_{next}$ values. Consider a replay buffer comprised of $[o, a, r, o_{t+1}]$, where $o$ is the observation, $a$ is the action, $r$ is the reward, and $o_{t+1}$ is the next observation. During the exploration phase, $o$ is fed into the actor network, which will output the action $a$. Subsequently, the actions are used as inputs to the critic network, along with the observation from the replay buffer, in order to obtain the Q-value. The training of the actor network strives to maximize Q. While initially, the outputs will be quite random, the actions become more sensible as training continues. To train the critic network, the Q-value obtained from directly using $o$ and $a$ are compared with the $Q_{next}$ values, which utilize the $o_{next}$ as the input. To prevent $Q$ and $Q_{next}$ from updating from the same critic network, a target network is constructed separated by the dashed line in the figure below, which freezes the critic network when training the actor network to obtain the Q-value, and both networks are frozen to obtain the $Q_{next}$ values. Ultimately, convergence is attained as $|Q - (r + \delta \cdot Q_{next})|$ is minimized, where $r$ is the cumulative reward and $\delta$ is the discount factor. The full algorithm is presented as Algorithm 1.

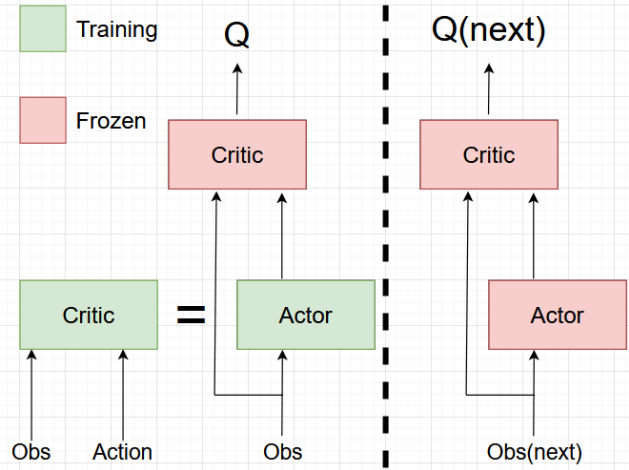

**Figure 3: DDPG Model Architecture for Obtaining Q and Q(next) Values**

**Model Overview**

The features for any given node $i$ include the raw velocity and position information. Each CAV with a sensing range $\rho$ can detect HDVs within the radius $\rho$, and the sensed data is updated at each time step. The assumption is that the CAVs are able to perform sensed data collection via onboard sensors such as LiDAR and computer vision. The collected information is used to construct a quadruplet $x_i$, which concatenates the velocities and positions of all vehicles within $\rho$. $x_i$ can then be shared among all CAVs in the system with latency $\gamma(d)$, which is a function dependent on the Euclidian distance between the information sender and receiver. For this study, the information is relayed to the centralized controller to concatenate raw information as a node features $X_t = [x_i]_{i=1}^{N}$. Further, the adjacency matrix $A$, as described previously, is utilized to preserve the graphical network topology.

As shown in Figure 2, the model structure is comprised of the following components: a Fully Connected Network (FCN) encoder, a GCN layer, the actor network, and the critic. The concatenated raw information, $X_t$, is sent into the FCN encoder $\varphi$ to generate the node embeddings $\mathcal{H} \subset R^{N \times d}$. At each time step $t$,

$$H_t = \varphi(X_t) \in \mathcal{H}$$

Subsequently, graphic convolution is performed on each vehicle in the embedding space $\mathcal{H}$. The GCN layer computes the node embeddings using the corresponding node embeddings as well as neighboring node embeddings, which is critical for using both CAV and HDV information for decision making. Further, it is important to note that this the dynamic length inputs resulting from the two data sources are resolved via the GCN layer. The embeddings are computed as follows:

$$Z_t = g(H_t, A_t) = \sigma(\widehat{D}_t^{-1/2} \hat{A}_t \widehat{D}_t^{-1/2} H_t W + b)$$

where $\hat{A}_t = A_t + I_N$ is the adjacency matrix, $\widehat{D}_t$ is the degree matrix computed for $\hat{A}_t$, and $\sigma$ is the activation function, namely, ReLU. After the fusion block, the node embeddings for the $Z_t^{CAV}$ CAVs are filtered out as the ingredient for the reinforcement learning module.

As stated previously, the reinforcement learning algorithm used in this paper is the DDPG agent. DDPG agent has 2 parts of neural networks, namely actor and critic (Lillicrap et al., 2016). In this study, both Actor and Critic are fully connected networks (FCNs). The Actor network output individual acceleration for each CAV $a_t$ directly from the node embeddings $Z_t^{CAV}$ for CAVs. Then the $Z_t^{CAV}$ and individual actions (accelerations) for all the CAVs $a_t$ are concatenated and flattened to feed into the Critic network.

The FCN structures for each component of the model are:

- FCN Encoder $\varphi$: $Dense(32) + Dense(32)$
- GCN layer $g$: $GraphConv(32)$
- Actor network $\mu$: $Dense(32) + Dense(32) + Dense(1)$
- Critic: $Dense(32) + Dense(1)$

With the exception of the output layer which uses linear activation, the activation functions for each block are all ReLU as stated previously. Since DDPG is an off-policy algorithm, an experience replay buffer R is implemented to store the transitions, and the model is trained by the random mini-batches sampled from the replay buffer R. The full algorithm including training is listed in Algorithm 1

---

**Algorithm 1** DDPG algorithm

Randomly initialize critic network $Q(s, a|\theta^Q)$ and actor $\mu(s|\theta^\mu)$ with weights $\theta^Q$ and $\theta^\mu$.
Initialize target network $Q'$ and $\mu'$ with weights $\theta^{Q'} \leftarrow \theta^Q$, $\theta^{\mu'} \leftarrow \theta^\mu$
Initialize replay buffer $R$
**for** episode = 1, M **do**
 Initialize a random process $\mathcal{N}$ for action exploration
 Receive initial observation state $s_1$
 **for** t = 1, T **do**
  Select action $a_t = \mu(s_t|\theta^\mu) + \mathcal{N}_t$ according to the current policy and exploration noise
  Execute action $a_t$ and observe reward $r_t$ and observe new state $s_{t+1}$
  Store transition $(s_t, a_t, r_t, s_{t+1})$ in $R$
  Sample a random minibatch of $N$ transitions $(s_i, a_i, r_i, s_{i+1})$ from $R$
  Set $y_i = r_i + \gamma Q'(s_{i+1}, \mu'(s_{i+1}|\theta^{\mu'})|\theta^{Q'})$
  Update critic by minimizing the loss: $L = \frac{1}{N}\sum_i (y_i - Q(s_i, a_i|\theta^Q))^2$
  Update the actor policy using the sampled policy gradient:

$$\nabla_{\theta^\mu} J \approx \frac{1}{N}\sum_i \nabla_a Q(s, a|\theta^Q)|_{s=s_i, a=\mu(s_i)} \nabla_{\theta^\mu} \mu(s|\theta^\mu)|_{s_i}$$

  Update the target networks:

$$\theta^{Q'} \leftarrow \tau\theta^Q + (1-\tau)\theta^{Q'}$$
$$\theta^{\mu'} \leftarrow \tau\theta^\mu + (1-\tau)\theta^{\mu'}$$

 **end for**
**end for**

---

**Experiment Settings**

To perform the RL experiment, this study utilized an open-source traffic simulator SUMO combined with a Python API that serves as a bridge between SUMO and reinforcement learning libraries (Krajzewicz et al., 2012; Wu et al., 2017).

Two highway bottleneck networks were constructed for simulation, training, and testing of the proposed method. The first simulated highway bottleneck network represents a moderately congested scenario without severe backflows affecting upstream traffic. In the second network, a more aggressive flow breakdown is simulated, with severe congestion approaching upstream due to the bottleneck. The moderately congested bottleneck network is shown in Figure 4. The 4-lane highway segment has a total length of 0.5 km with a lane drop to 3 lanes at the 0.3 km mark and another lane drop at the 0.4 km mark. The lane drops occur only at the rightmost lane. Further, the inflows of vehicles into the system are set as 1500 $vehs/h$, and the market penetration of CAVs is set at 10%. The total number of vehicles in the network for each episode is 50, with 5 CAVs and 45 HDVs. Lastly, the information latency $\gamma(d)$ is assumed to be zero.

For the vehicle parameters, the lane change model LC2013 (Erdmann, 2015) was utilized for both HDVs and CAVs. The longitudinal control for HDVs follows the Intelligent Driver Model (Treiber et al., 2000), and that of the CAVs follow the proposed DRL model.

Prior to training, a warm-up phase is defined, and the CAVs perform relatively random actions as the replay buffer is filled with random transitions. In this study, both the actor and critic receive a warm-up phase of 1000 steps. Once the training begins, the transition batches with batch size of 64 were sampled randomly. The overall training horizion was defined as $10^6$ steps (approximately 800 epochs). *ADAM*, a method for stochastic optimization (Kingma and Ba, 2015), was utilized for the optimization parameters with an initial learning rate of $\gamma = [10^{-4}, 10^{-3}]$ and a soft target model update rate $\tau = 10^{-3}$.

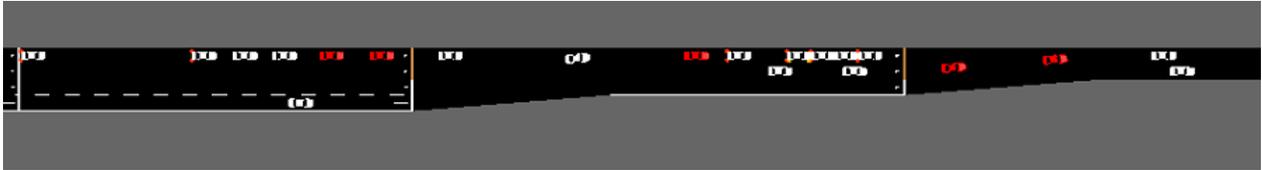

**Figure 4: Moderately Congested Bottleneck Network**

To replicate a more severely congested highway bottleneck, a second network was constructed as shown in Figure 5. This network is a 1-km long corridor comprised of 4-lanes along the first 0.6 km, followed by a 2-lane edge for the next 0.2 km and a 1-lane edge for the final 0.2 km. The inflows of vehicles are set as 2300 $vehs/h$, with 140 total vehicles in the system per episode, with 10 CAVs and 130 HDVs, thereby possessing fewer percent share of agents in the traffic stream than the moderately congested scenario. All other parameters are identical to the first network.

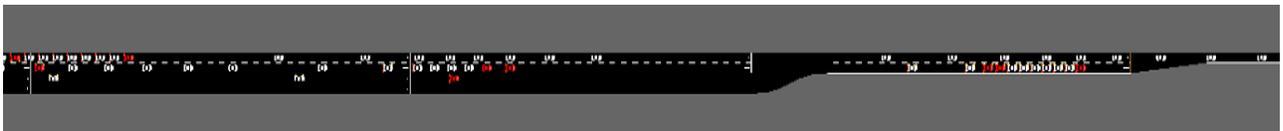

**Figure 5: Severely Congested Bottleneck Network**

1) State Space

At each time step $t$, $N$ CAVs are available within the training environment. The state space is comprised of the node features ($X_t$) and the adjacency matrix ($A_t$). The node features contain data regarding vehicular velocities ($v_i$) and positions ($p_i$). The adjacency matrix is an $N \times N$ matrix with binary entries that correspond to the network topology. $A_{ij} = 1$ when vehicle $i$ and $j$ are connected, and $A_{ij} = 0$ otherwise.

2) Action Space

At each time step $t$, a given CAV can perform as its action a change in its maximum acceleration (or deceleration). The upper bounds for acceleration and deceleration are defined as $3\ m^2/s$ and $-3\ m^2/s$, respectively. Therefore, the action is a real value $a_i \in [-3,3]$, and the action space for the centralized controller is $\mathcal{A} = \{a_i\}_{i=1}^N$. On the other hand, the acceleration of HDVs in the system are governed by the Intelligent Driver Model, which calculates the acceleration based on headway $s_a$, velocity $v_a$, and relative velocity $\Delta v_a$, and desired headway $s^*$ (Treiber et al., 2000):

$$a_{HDV} = \frac{dv_{HDV}}{dt} = a \left[ 1 - \left( \frac{v_a}{v_0} \right)^\delta - \left( \frac{s^*(v_a, \Delta v_a)}{s_a} \right)^2 \right]$$

3) Reward Function

The reward function utilized in this RL experiment is comprised of two components: a reward for throughput and a penalty for high speed variance. The throughput reward is defined as the number of vehicles that have left the highway segment over the past few time steps:

$$r_1 = 3600 \sum_{i=t-T}^{t} \frac{n_{exit}}{T}$$

The penalty for high speed variance is a scalar multiple of the speed variance between vehicles prior to the lane drops:

$$r_2 = -\frac{\beta \sigma^2}{n_\mu}$$

where $\beta$ is a constant parameter, $\sigma^2$ is the variance, and $n_\mu$ is the number of vehicles which have not yet entered the bottleneck. As the primary idea behind speed harmonization is to have vehicles operate at low speed variances, penalizing high speed variance encourages the CAV to control local traffic to operate with harmonized speeds. Combining the penalty term and the reward term, the total reward is calculated as:

$$R = r_1 + r_2 = 3600 \sum_{i=t-T}^{t} \frac{n_{exit}}{T} - \frac{\beta \sigma^2}{n_\mu}$$

**RESULTS**

*Moderately Congested Scenario*

Figure 6 presents the reward curve over 800 epochs ($10^6$ time steps) of the training cycle for the moderately congested scenario. The first 200 epochs serve as the warm-up, which allow the model to avoid early overfitting. After the warm-up, the training ultimate converges after 400 epochs. The reward obtained from the rule-based model is a fixed value, as no training occurs. Without training, the reward output cannot be improved upon. Compared to the rule-based model, which exhibits the IDM car following behavior, the DRL-based approach is superior in both stabilizing speed deviations as well as increasing throughput at the end of the bottleneck segment.

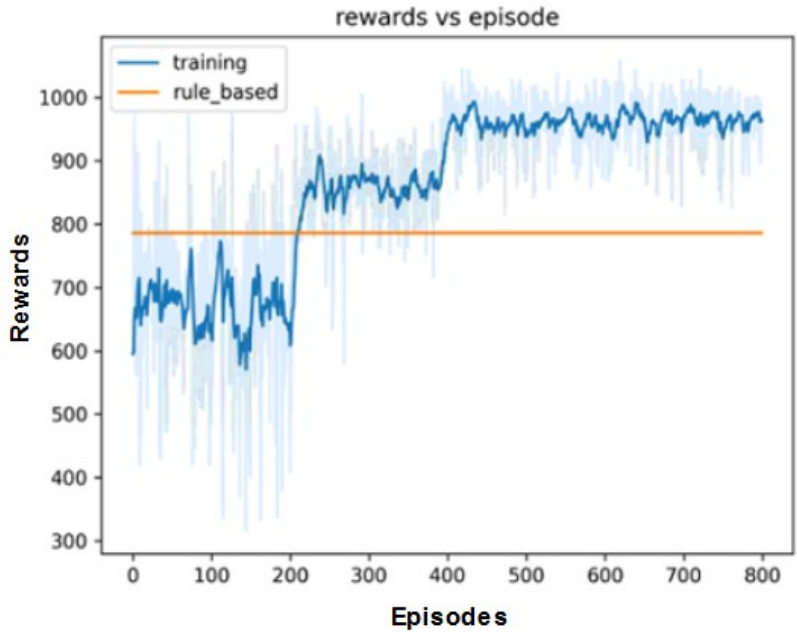

**Figure 6: Reward Curve Comparing Rule-based and RL-based Controllers for Moderate Congestion**

The rendered policy shows that the CAVs utilize two strategies to mitigate bottleneck congestion. If a CAV is on a lane that is dropped downstream, it increases its speed to merge towards the left. On the other hand, if a CAV is on the two leftmost lanes which are not dropped in the bottleneck, the CAV appropriately slows its speed in order to create gaps for HDVs to merge in. The CAVs' role is similar to the traffic stabilizing vehicle (Kreidieh et al., 2018; Stern et al., 2018) as well as mimicking a "generous" human driver who allows vehicles to cut in front.

The time-space heatmap of traffic speeds for the trained controller and rule-based controller are presented in Figure 7 and Figure 8, respectively. In these plots, the x-axis is the time of the episode, and the y-axis is the spatial representation of the network. The color in each cell represents the average speeds for a given space-time of 50 m and 10 seconds. Therefore, the top-left cell would indicate the start of the segment at the beginning of an episode, the bottom left-cell indicates the end of the segment at the beginning of the episode, and so on. In both plots, the bottleneck (the large, dark-red region) is clearly visible. The key difference is that, under the trained controller, the duration and intensity of congestion is shorter than that of the rule-based controller. As such, the episode duration is shorter as well, with the rule-based episode lasting 300 more time-steps. This signifies an improvement in mobility, allowing vehicles to remain congested for a shorter duration.

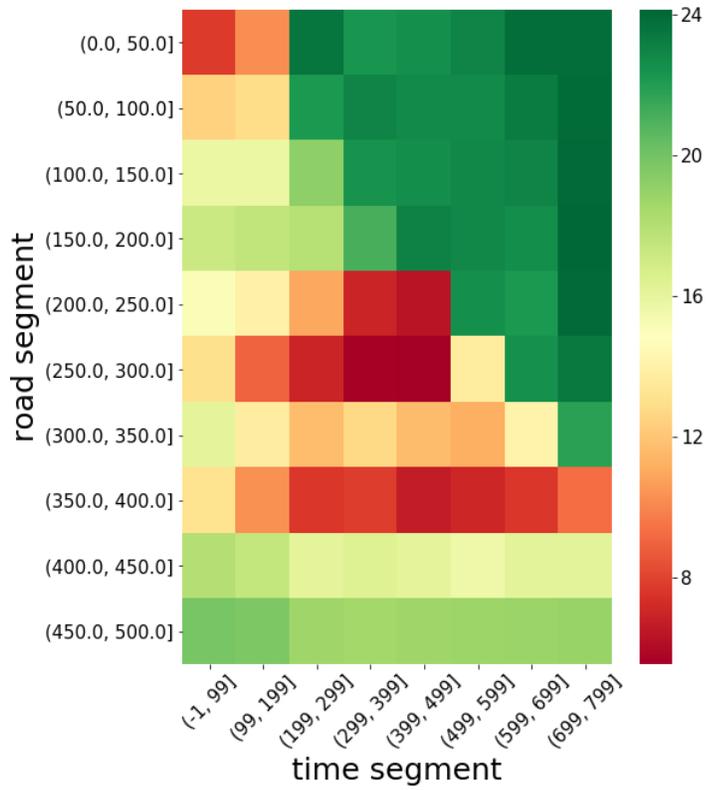

**Figure 7: Time-Space Heatmaps of Mean Traffic Speeds using RL-based Controller for Moderate Congestion**

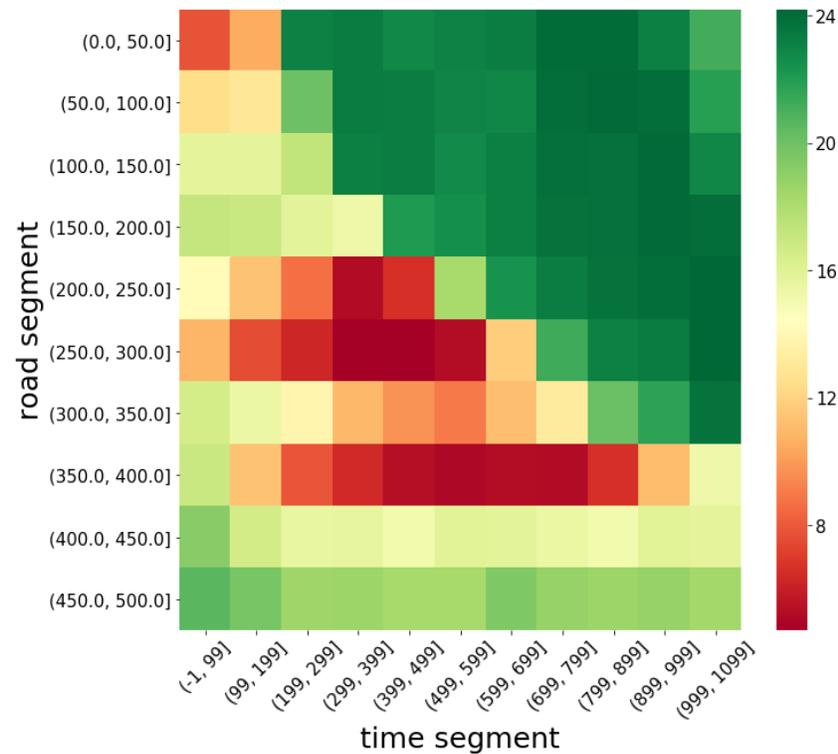

**Figure 8: Time-Space Heatmape of Mean Traffic Speeds using Rule-based Controller for Moderate Congestion**

***Severely Congested Scenario***
In the severely congested scenario, the time horizon is defomed as 1500 time-steps. The reward curve for the severely congested scenario is presented in Figure 9. It can be observed that the RL-based controller outperforms a rule-based controller, though not as significantly as in the moderately congested scenario.

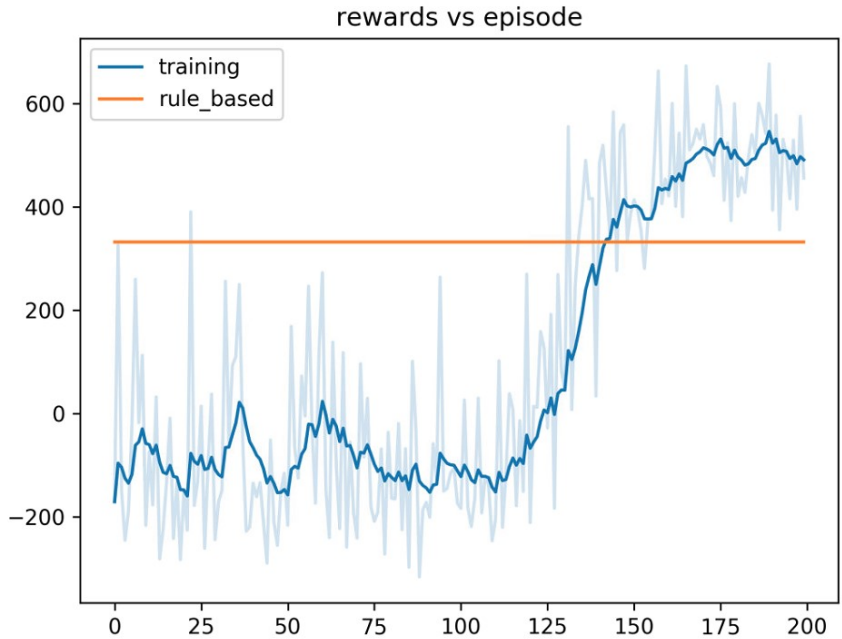

**Figure 9: Reward Curve Comparing Rule-Based and RL-based Controllers for Severe Congestion**

Unlike the moderately congested scenario, no episode was terminated before the defined time horizon as shown in Figures 10 and 11, which indicates that the RL-based controller is unable to induce all HDVs to exit the bottleneck within the defined time horizon of 1500 time-steps.

Additionally, the traffic patterns are significantly different in the RL-based and rule-based controllers. Figure 10 shows two areas of slow-moving traffic for the RL-based controller: in the 200 m – 400 m segment and the 650 m – 800 m segment. On the other hand, Figure 11 shows that the rule-based controller has one area of slow-moving traffic: in the 500 m – 800 m segment. Considering that there are successive lane drops at 600 m and 700 m, the rule-based controller causes severe congestion immediately prior to the bottleneck. On the other hand, the RL-based controller improves throughput by controlling the inflows to the bottleneck region. The slowdown area from 200 m – 400 m correspond to the inflow control strategy that the CAVs utilize to reduce the rate of vehicles entering the successive lane drops. As a result, traffic backflows can be reduced, and flow conditions can be improved in the bottleneck region as is shown by the significantly smaller area of low mean speeds at the bottleneck.

Another important metric for consideration is the speed harmonization, that is, the reduction of speed standard deviation (SSD) prior to the bottleneck. Exhibiting high SSD prior to the bottleneck could indicate traffic shockwave formation, which can propagate over time to affect beyond the localized region of the bottleneck. Figure 12 shows the time-space heatmap of SSD using the RL-based controller, and Figure 13 shows that of the rule-based controller. The region of interest is the section immediately prior to and containing the bottleneck, which is the 400 m – 800 m segment shown in the bottom left quadrant of the heatmap plots shown in Figures 12 and 13. Since the RL-based controller reduces congestion by controlling the inflows into the bottleneck region, the SSD in the bottom left quadrant of Figure 12 shows a slight improvement with respect to the affected time-space cells compared to that of the rule-based controller shown in Figure 13. However, this is achieved by slowing down the vehicles upstream, which is the reason why the upper left and right quadrants show high SSDs.

Further, Figure 14 presents a comparison of throughput between the trained RL-based and rule-based controllers. Over 10 episodes, it can be seen that the RL-based model consecutively has a higher throughput than the rule-based model.

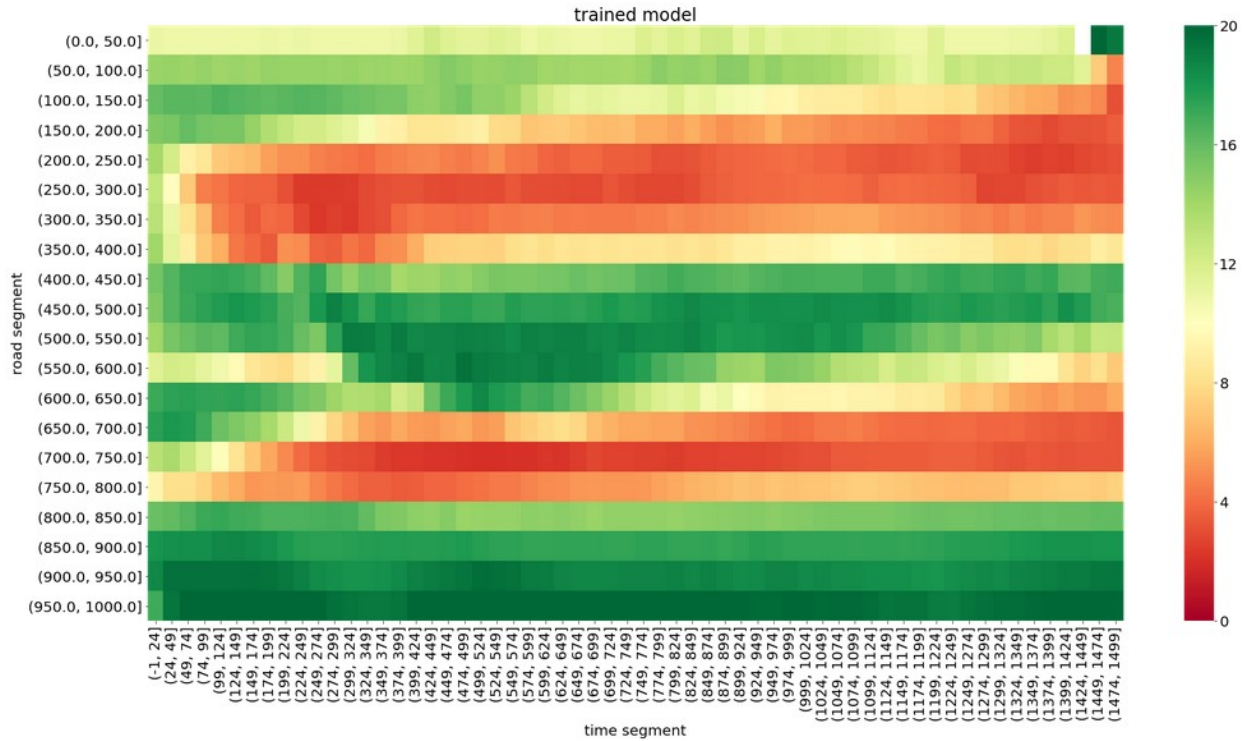

**Figure 10: Time-Space Heatmaps of Mean Traffic Speeds using RL-based Controller for Severe Congestion**

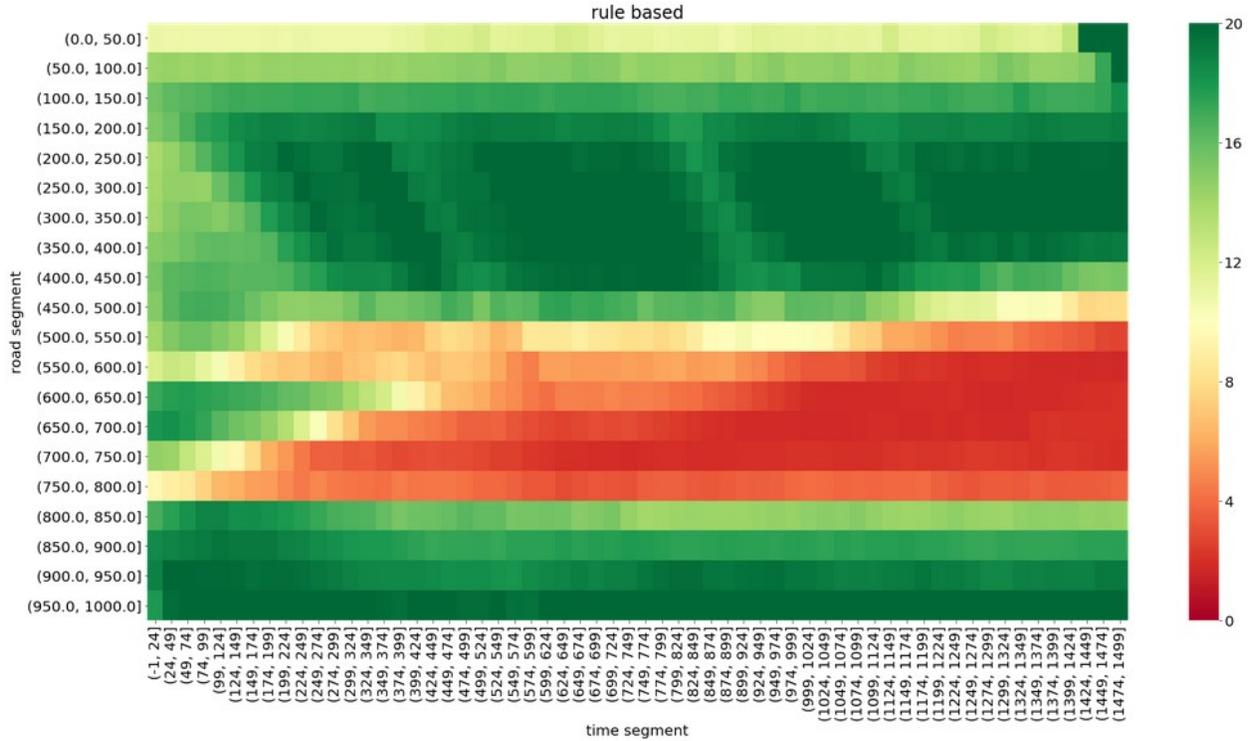

**Figure 11: Time-Space Heatmaps of Mean Traffic Speeds using Rule-based Controller for Severe Congestion**

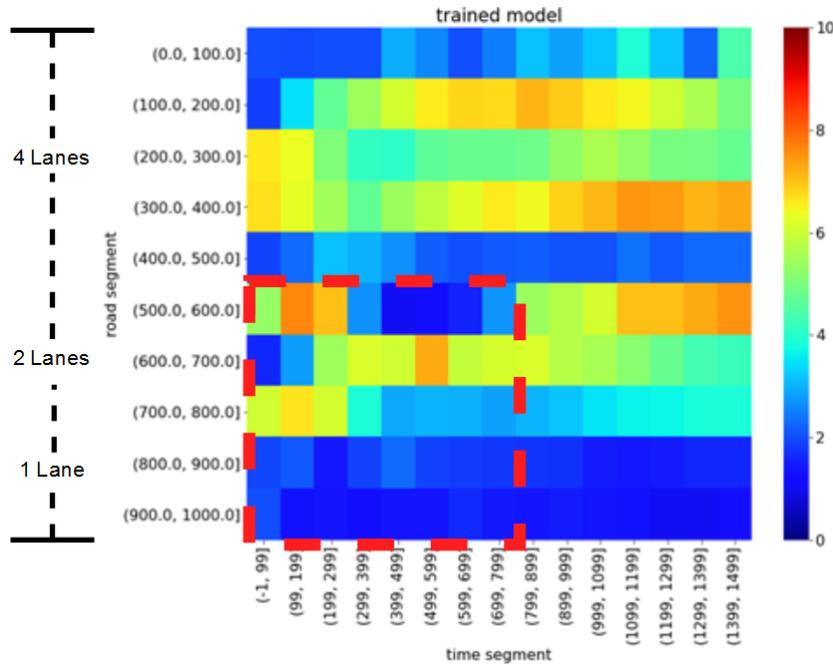

**Figure 12: Time-Space Heatmaps of Speed Standard Deviations using RL-based controller for Severe Congestion**

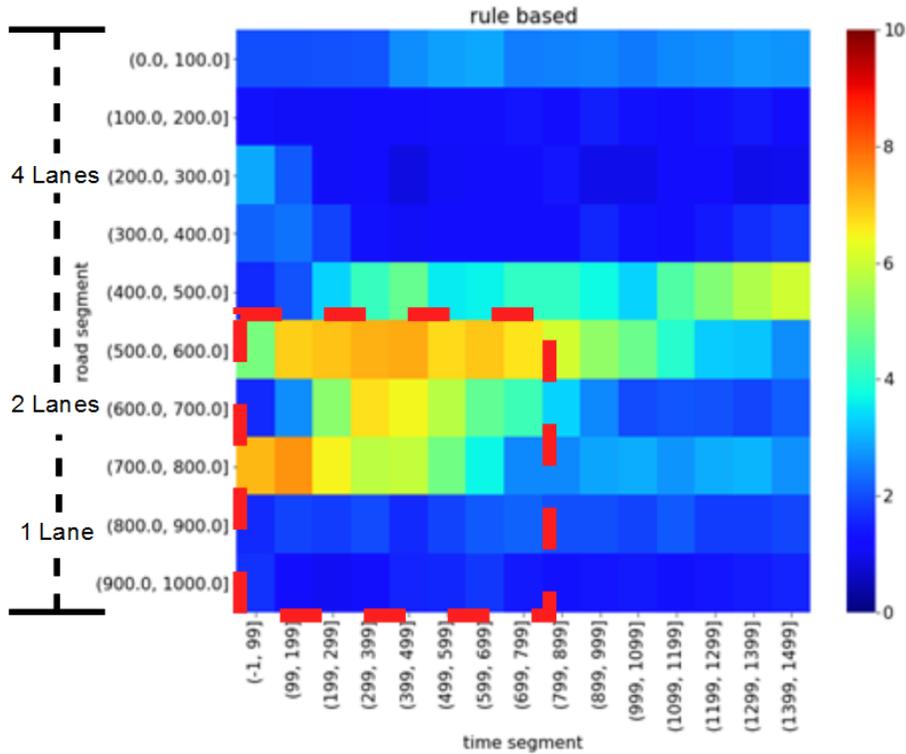

**Figure 13: Time-Space Heatmaps of Speed Standard Deviations using Rule-based controller for Severe Congestion**

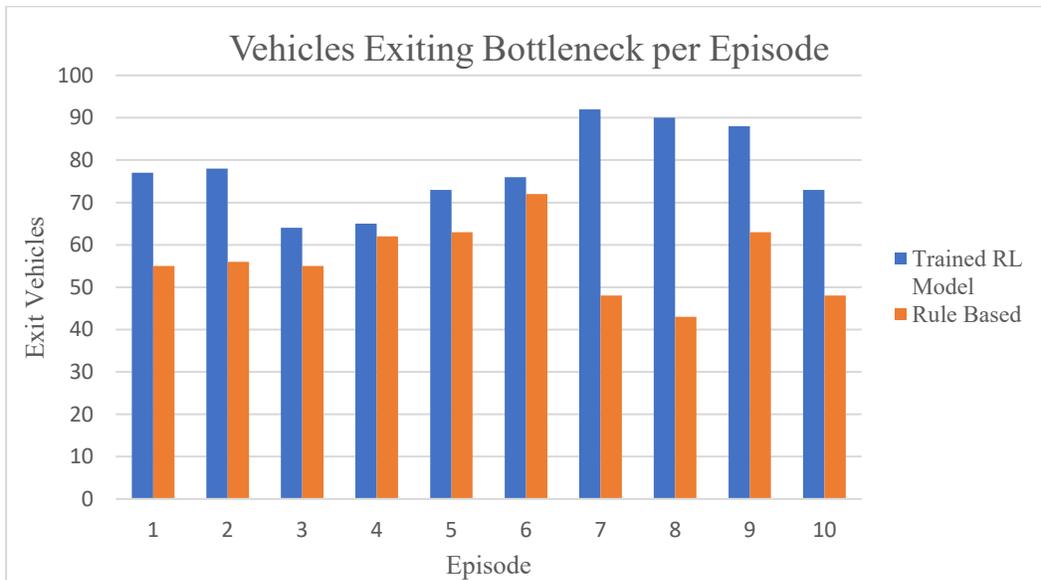

**Figure 14: Vehicles Exiting Bottleneck per Episode using RL-based and Rule-based Controllers**

**CONCLUDING REMARKS**

This study applied reinforcement learning algorithms to train CAVs the driving behavior that can be utilized for mitigating highway bottleneck congestion. The challenge to RL in highway environments is the dynamic number of vehicles and continuous vehicle control parameters. To address the continuous actions required for CAV control, the Deep Deterministic Policy Gradient algorithm was used to train the RL agents. Due to the dynamic length inputs caused by the highly varying number of vehicles observed, the Graphical Convolutional Network was utilized. This method was applied to highway congestion mitigation scenarios, particularly those caused by physical highway bottlenecks from successive lane drops. Multiple CAVs were trained to perform collaborative maneuvers to improve the throughput as well as to reduce the speed deviations in the traffic stream prior to the bottlenecks.

Based on the experiment settings, the CAVs collaboratively influenced the driving of the HDVs by altering the speeds of lanes that cause high speed variations in the segment. The findings of this study present a strong evidence for the use of learning-based CAV controllers that not only assists in ego vehicle operations but also serves as a tool for safety and mobility improvements for a localized network. As shown in Figure 6 and Figure 9, the reward output using the RL-based controller outperforms that of the rule-based controller for both the moderately and severely congested scenarios. Further, in the highly congested scenario, the throughput is significantly increased using the RL-based controller, as shown in Figure 14. Thus, there is evidence to conclude that the RL-based controller can be successful in mitigating highway bottleneck congestion in both moderately and severely congested scenarios.

The behavior of CAVs in moderately congested and severely congested scenarios are slightly different. In the moderately congested scenario, CAVs operate to create gaps for HDVs in dropped lanes to merge into so that queuing is reduced at the bottleneck. They act as cooperative vehicles willing to sacrifice its own travel time to improve the network flow. The improvements can be observed in Figure 7 and Figure 8, which show that the congestion at the bottleneck is both less severe and shorter in duration when using the RL-based controller. Further, the number of space-time cells with average speeds lower than 8 km/h was 10 for the RL-based controller, whereas that of the rule-based controller was 15. This is further evidence that the severity and duration of congestion is of a lesser degree with the RL-based controller.

In the severely congested scenario, CAVs can no longer find great benefit in simply allowing HDVs to merge out of dropped lanes, as there are too many vehicles in the dropped lanes. Instead, mitigation strategies take place much higher upstream. CAVs in the 100 m – 400 m segment travel at low speeds to control the inflows into the bottleneck region, subsequently reducing queueing. As a result, regions of low mean speeds and high SSDs appear prior to the bottleneck. However, the resulting outflow is still improved.

Given such results, combining GCN and DDPG to train CAVs for congestion mitigation is largely successful in reducing congestion from physical bottlenecks. Some open questions not addressed in this study include the use of lane changing as an action for CAVs. Literature has shown that lane-level traffic management can be highly effective in congestion mitigation (Nagalur Subraveti et al., 2019; Vinitsky et al., 2018). Thus, utilizing CAV lane occupancy as an action may be more effective in inducing HDV lane change behavior but possibly at the cost of crash potential.

**ACKNOWLEDGMENTS**

This work was supported by Purdue University's Center for Connected and Automated Transportation (CCAT), a part of the larger CCAT consortium, a USDOT Region 5 University Transportation Center funded by the U.S. Department of Transportation, Award #69A3551747105. The contents of this paper reflect the views of the authors, who are responsible for the facts and the accuracy of the data presented herein, and do not necessarily reflect the official views or policies of the sponsoring organization.

**AUTHOR CONTRIBUTIONS**

The authors confirm contribution to the paper as follows: all authors contributed to all sections. All authors reviewed the results and approved the final version of the manuscript.